\documentclass[sigconf, screen]{acmart}
\AtBeginDocument{%
  }

\copyrightyear{2026}
\acmYear{2026}
\setcopyright{acmlicensed}
\acmConference[MM '26] {Proceedings of the 34rd ACM International Conference on Multimedia}{November 10--14, 2026}{Rio de Janeiro, Brazil}
\acmBooktitle{Proceedings of the 34rd ACM International Conference on Multimedia (MM '26), November 10--14, 2026, Rio de Janeiro, Brazil}
\usepackage{placeins}
\usepackage{multirow}
\usepackage{graphicx}\usepackage{subcaption}

\begin{document}

\title{Test-Time Adaptation via Dual Distillation for Videos Under Severe Distribution Shifts}

\author{André Sacilotti}
\email{andre.sacilotti@alumni.usp.br}
\orcid{0000-0001-9359-4298}
\affiliation{%
  \institution{University of São Paulo}
  \city{São Carlos}
  \state{São Paulo}
  \country{Brazil}
}

\author{Samuel Felipe dos Santos}
\email{samuel.felipe@ufscar.br}
\orcid{0000-0001-6061-5582}
\affiliation{%
  \institution{Federal University of São Carlos}
  \city{Sorocaba}
  \state{São Paulo}
  \country{Brazil}
}

\author{Jurandy Almeida}
\email{jurandy.almeida@ufscar.br}
\orcid{0000-0002-4998-6996}
\affiliation{%
  \institution{Federal University of São Carlos}
  \city{Sorocaba}
  \state{São Paulo}
  \country{Brazil}
}

\renewcommand{\shortauthors}{Sacilotti, Santos, and Almeida}

\begin{abstract}
Deep learning models have achieved state-of-the-art performance in several computer vision tasks. However, they experience severe performance degradation when applied to real-world scenarios due to unanticipated distribution shifts. Test-Time Adaptation (TTA) attempts to solve this problem by using unlabeled data from the target domain to dynamically adapt to the test distribution at inference time, without access to the source data. However, TTA remains a challenging problem when adapting to continuous, temporally correlated data, such as videos, and in scenarios where the target domain contains severe domain shifts. For this reason, few works in the literature explore TTA for videos under such extreme conditions. To overcome these limitations, we propose \textbf{Test-time Adaptation via Dual Distillation (TADD)}, an online adaptation framework that relies on a lightweight projection adapter to bridge the domain gap. The adapter module is pre-trained on the source domain and then adapted to the target using our proposed complementary losses: (\textit{i}) \emph{zero-shot distillation}, which encourages alignment with the domain-agnostic features from a pre-trained vision-language model~(VLM); and (\textit{ii}) \emph{target distillation}, which retains the source domain discriminative knowledge encoded in the pre-trained adapter. Built upon a frozen CLIP backbone, our method introduces this lightweight projection adapter as the sole updatable component during inference. We conducted extensive evaluations on three well-known video action recognition benchmarks: UCF-HMDB, Daily-DA, and Sports-DA. Our experiments in the closed-set scenario demonstrate that our method consistently outperforms state-of-the-art TTA baselines. Notably, our TTA approach improves upon previous methods by up to +3.81\% on UCF-HMDB, +2.63\% on Daily-DA, and +3.03\% on Sports-DA.
\end{abstract}

\ccsdesc[500]{Computing methodologies~Transfer learning}
\ccsdesc[300]{Computing methodologies~Action recognition}

\keywords{Test-Time Adaptation, Video Action Recognition, Domain Adaptation, Vision-Language Model}

\maketitle

\section{Introduction}
Deep learning models, and particularly the recent wave of Vision-Language Models~(VLMs), have achieved unprecedented performance across a wide array of visual and language tasks~\cite{li2025survey}. However, this success heavily relies on the assumption that the training and testing data share an identical distribution~\cite{liang2025comprehensive}. When deployed in the wild, these models frequently encounter unanticipated distribution gaps---such as those caused by varying lighting conditions, weather effects, sensor degradation, or lossy data compression. These shifts cause severe performance degradation, critically hindering the reliability of these models in practical applications~\cite{liang2025comprehensive}.

To mitigate the impact of distribution shifts, several transfer learning paradigms have emerged. Unsupervised Domain Adaptation~(UDA) aligns a model trained on a labeled source domain with a target domain by adapting it with labeled source and unlabeled target data~\cite{liang2025comprehensive}. However, UDA requires access to source data during the adaptation stage, which is often infeasible due to strict privacy restrictions or impractical due to the data being stored in a decentralized manner~\cite{liang2020we}.

Source-Free Domain Adaptation~(SFDA) is a more challenging paradigm that adapts a source-trained model to the target domain without any access to the original source data~\cite{zara2023unreasonable}.
However, SFDA methods typically operate offline, where the adaptation requires multiple iterative passes over the target dataset prior to inference. Test-Time Adaptation~(TTA) extends the principles of SFDA into a strictly online setting, dynamically adjusting the pre-trained model to each new batch of the continuous stream of unlabeled test data during inference~\cite{xiao2024beyond}.

While TTA has seen rapid advancement in the image domain, extending it to videos introduces significant difficulties. Video streams exhibit complex temporal dynamics and high inter-frame correlation~\cite{lin2023video}. Consequently, standard image-based adaptation strategies, particularly those based on entropy minimization, suffer from severe performance drops, model degeneration, and catastrophic forgetting when dealing with temporally correlated features and continual distribution shifts~\cite{fahim2024st2st}. Furthermore, the limited existing literature on video TTA predominantly evaluates methods against synthetically added noise (e.g., simulated illumination changes or motion blur) rather than actual cross-domain benchmarks featuring significant, natural distribution gaps.

To overcome these critical limitations, we propose \textbf{Test-time Adaptation via Dual Distillation (TADD)}, a fully online adaptation framework tailored for videos under severe, real-world distribution shifts. Our approach leverages the robust priors of foundation VLMs to anchor the adaptation process. Built upon a frozen CLIP~\cite{radford2021learning} backbone, our method utilizes a lightweight projection adapter as the sole updatable component during inference.

To optimize this target adapter without catastrophic forgetting, we introduce a pair of complementary distillation losses: (\textit{i}) the \textbf{zero-shot distillation} loss, which aligns the adapter's predictions with the frozen CLIP's zero-shot predictions, preserving domain-agnostic vision-language knowledge and preventing semantic collapse; and (\textit{ii}) the \textbf{target distillation} loss, which distills the source domain's discriminative knowledge, encoded within a frozen, pre-trained source adapter, directly to the target adapter.

We conduct extensive experiments on three challenging video action recognition benchmarks under natural domain shift: UCF-HMDB, Daily-DA, and Sports-DA. Our experiments show that TADD consistently outperforms state-of-the-art TTA baselines. Furthermore, we demonstrate that our method maintains exceptional robustness and high accuracy in more challenging classification scenarios, including partial-set and multi-target domain adaptation.

In summary, our main contributions are:
\begin{itemize}

    \item  We introduce TADD, a novel test-time adaptation framework for video action recognition designed to address severe, natural distribution shifts rather than synthetic corruptions.

    \item We propose a complementary dual distillation loss that successfully balances alignment with a domain-agnostic vision-language model (via zero-shot distillation) alongside source-specific discriminative knowledge (via target distillation).

    \item We establish state-of-the-art test-time adaptation performance across multiple realistic video TTA benchmarks, improving upon previous methods by up to +3.81\% on UCF-HMDB, +2.63\% on Daily-DA, and +3.03\% on Sports-DA.
\end{itemize}

The remainder of this paper is organized as follows. Section~\ref{sec:related_works} reviews related work in source-free domain adaptation and test-time adaptation. Section~\ref{sec:our_approach} details our proposed TADD framework, outlining the problem definition, architecture, and the dual distillation objective. Section~\ref{sec:experiments} presents the experimental setup, benchmark results, and comprehensive ablation studies. Finally, Section~\ref{sec:conclusions} concludes the paper and discusses future directions.

\section{Related Work}
\label{sec:related_works}

In real-world deployments, deep learning models frequently encounter unanticipated distribution shifts that severely degrade their performance~\cite{liang2025comprehensive}.
SFDA mitigates this by aligning source and target distributions without requiring access to the original source data~\cite{zara2023unreasonable}.
For instance, in the image domain, SHOT~\cite{liang2020we} tackles SFDA by freezing the source classifier and training target-specific feature extractors via information maximization and self-supervised pseudo-labeling.
In the video domain, ATCoN~\cite{xu2022atcon} addresses SFDA by enforcing temporal consistency across video clips and attentively aggregating them based on their confidence relative to a frozen source classifier.
EXTERN~\cite{xu2024extern} optimizes temporal features using endo- and exo-temporal regularizations, relying solely on the output predictions from a black-box source model, a more restrictive scenario, known as black-box SFDA, where models are deployed as opaque APIs.
DALL-V~\cite{zara2023unreasonable} leverages VLMs by training separate source and target adapters on a frozen CLIP backbone, subsequently distilling the knowledge from both adapters and the frozen backbone into a student network. 
While SFDA successfully eliminates the need for source data, it remains an offline, iterative fine-tuning process that must be completed prior to inference.

TTA extends the principles of SFDA into a strictly online setting, achieving model adaptation during or alongside inference.
This online paradigm reduces computational cost and improves generalization~\cite{xiao2024beyond}.
In the image domain, TENT~\cite{wang2020tent} minimizes the entropy of the predictions by estimating normalization statistics and updating channel-wise affine transformations in each batch.
WATT~\cite{osowiechi2024watt} adapts the visual encoder of CLIP with pseudo-labels generated by a diverse set of text prompt templates, averaging the weights to consolidate global information.
BATCLIP~\cite{maharana2025batclip} proposes a bimodal approach that updates the parameters of normalization layers of both the visual and text encoders of CLIP, reinforcing multimodal alignment by associating image class prototypes and the text features.

Despite rapid advancements in image-based TTA, adapting video models at test time remains highly challenging. Video streams inherently contain complex temporal dynamics, high inter-frame correlation, and severe natural distribution shifts~\cite{lin2023video,fahim2024st2st}
Consequently, only a few methods have successfully tackled video TTA.
ViTTA~\cite{lin2023video} aligns the online statistical estimates of the test set with source training values, enforcing prediction consistency across multiple augmented views of a single test sample.
Alternatively, ST2ST~\cite{fahim2024st2st} introduces a self-distillation approach where the network undergoes an intra-video logit minimization phase, dynamically updating its predictions for temporally correlated incoming data.

Our proposed method, TADD, advances this nascent group of video TTA approaches. TADD introduces an efficient dual distillation framework on a frozen vision-language backbone. By anchoring the adaptation to a robust zero-shot prior while simultaneously distilling source-specific discriminative knowledge, TADD provides a stable, fully online solution tailored for severe distribution shifts in videos.

\begin{figure*}[t!]
  \centering
  \begin{subfigure}[t]{0.85\textwidth}
    \centering
    \includegraphics[width=\linewidth]{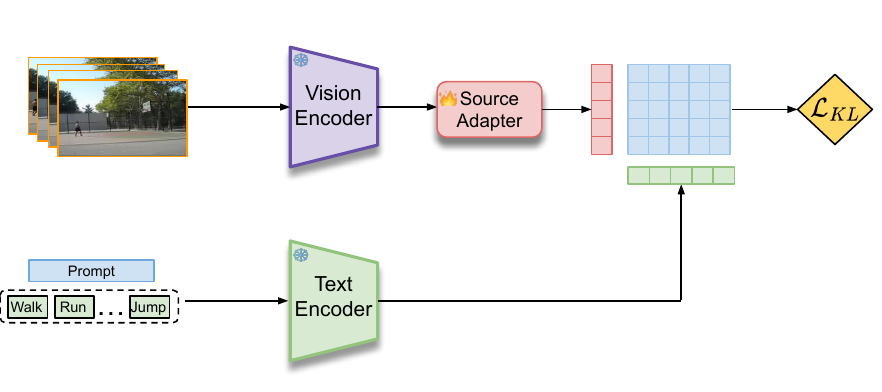}
    \caption{Source-domain pre-training of $G_{s}$ on labeled video, with CLIP's visual encoder frozen.}
    \label{fig:tadd-source}
  \end{subfigure}
  \\
  \begin{subfigure}[t]{\textwidth}
    \centering
    \includegraphics[width=\linewidth]{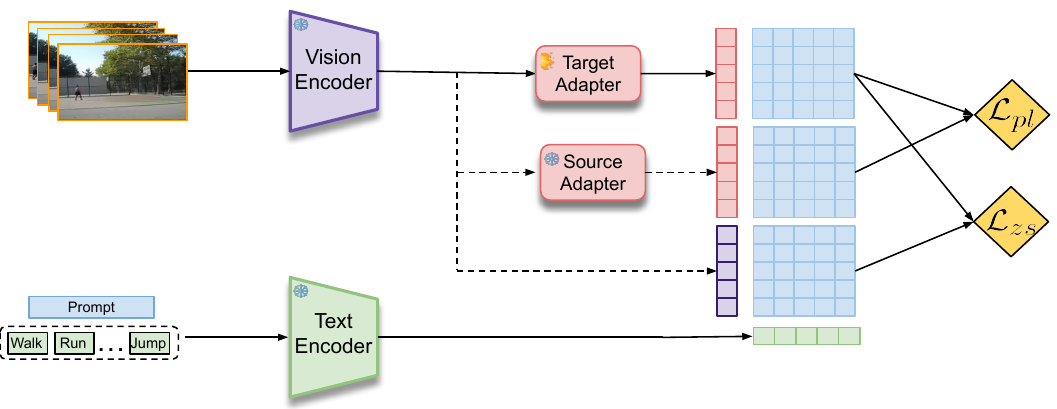}
    \caption{Online test-time adaptation with dual distillation, which updates $G_{t}$ on unlabeled target batches while $G_{s}$ and CLIP stay fixed.}
    \label{fig:tadd-tta}
  \end{subfigure}
  \caption{Overview of TADD. \textbf{(a)}~The source adapter $G_{s}$ is trained on the source domain. \textbf{(b)}~At test time, only the target adapter $G_{t}$ is updated via $\mathcal{L}_{\text{zs}}$ and $\mathcal{L}_{\text{pl}}$, and inference uses $G_{t}$ with frozen text embeddings.}
  \label{fig:tadd-overview}
\end{figure*}

\section{Our Approach}
\label{sec:our_approach}

In this section, we describe TTAD, our proposed method for tackling TTA in videos under severe distribution changes. 
Section~\ref{sec:problem} presents a formal definition of the video TTA problem.
Section~\ref{sec:background} details the architectural framework and how the VLM backbone is used for videos.
Section~\ref{sec:approach} introduces our dual distillation approach.
Finally, Section~\ref{sec:intutition} provides the core intuition driving this methodology.

\subsection{Problem Definition}
\label{sec:problem}

In this work, we study test-time adaptation for video action recognition under domain shift.
Formally, there is a source joint distribution $\mathcal{S} = p(x_s,y_s)$ and a target joint distribution $\mathcal{T} = p(x_t,y_t)$, where $(x,y)$ are pairs sampled from the joint space $\mathcal{X} \times \mathcal{Y}$. Here, $\mathcal{X}$ represents the input data feature space and $\mathcal{Y}$ represents the label space.

During the initial training phase, only data from $\mathcal{S}$ is available.
We assume a scenario characterized by a distribution gap between $\mathcal{S}$ and $\mathcal{T}$, such that $p(x_s,y_s) \neq p(x_t,y_t)$. 
Consequently, the predictions from a model trained solely on data from $\mathcal{S}$ can be highly unreliable when applied to data from $\mathcal{T}$.
To mitigate this, TTA adapts the model at test time while strictly having access only to unlabeled data $x_t$ from $\mathcal{T}$.
Furthermore, adaptation and inference must occur concurrently—either online or batch-by-batch—without accessing large amounts of target samples at any given step.

The ultimate goal is to improve recognition accuracy on the target distribution while keeping computational and storage overhead suitable for online deployment.

\subsection{Model Architecture and Initialization}
\label{sec:background}

To establish a robust anchor against domain shifts, we build our method upon a frozen VLM backbone, specifically, CLIP ViT-B/32.
Drawing inspiration from the DALL-V~\cite{zara2023unreasonable} framework for SFDA, we mitigate the domain gap without fine-tuning the CLIP backbone by placing a lightweight projection adapter on top of its visual encoder.

We utilize two distinct adapters in our pipeline: a source-pretrained adapter, denoted as $G_{s}$, and a target-updated adapter, denoted as $G_{t}$. 
The source adapter $G_{s}$ is trained entirely offline on labeled source videos and is subsequently frozen. 
Conversely, $G_{t}$ is the only module that receives gradient updates during TTA. 
At the onset of test-time adaptation, we initialize the weights of $G_{t}$ as a direct copy of the pre-trained $G_{s}$.
Following~\cite{gao2024clip}, we implement the adapters as two-layer MLPs with a ReLU activation function.

Since CLIP was originally developed for static images, we adapt it for video inputs $V \in \mathbb{R}^{T \times C \times H \times W}$ by encoding each frame using the frozen visual encoder $f$, and then averaging the resulting features over time to obtain a video-level representation $\mathbf{z}$:
\begin{equation}
    \mathbf{z} = \frac{1}{T} \sum_{t=1}^{T} \frac{f(V_t)}{| f(V_t) |}
\end{equation}

\noindent where $V_t$ denotes the $t$-th frame of the video. To obtain the classification logits, we embed $K$ class names with CLIP's frozen text encoder using prompts such as \textit{``a video of a [CLASS]''}, yielding a set of text embeddings $\{ \mathbf{w}_c \}_{c=1}^{K}$. The logits $\ell_c$ are defined as the scaled cosine similarities:
\begin{equation}
    \ell_c = \langle h(\mathbf{z}),\, \mathbf{w}_c \rangle
\end{equation}

\noindent where $h$ represents the identity function for zero-shot CLIP, the frozen source adapter $G_{s}$, or the updatable target adapter $G_{t}$.

\subsection{Test-time Adaptation via Dual Distillation}
\label{sec:approach}

Unlike DALL-V~\cite{zara2023unreasonable}, which relies on offline pseudo-labeling, TADD adapts $G_{t}$ dynamically by minimizing a dual distillation objective built from two Kullback-Leibler~(KL) divergence terms.
First, the \textbf{zero-shot distillation} loss matches the student distribution to CLIP's zero-shot predictions. 
This preserves the domain-agnostic vision-language alignment inherent to the VLM backbone:
\begin{equation}
  \mathcal{L}_{\text{zs}} = D_{\text{KL}}\!\Big( G_{t}(\mathbf{z}) \cdot \mathbf{W}^\top \;\Big\|\; \mathbf{z} \cdot \mathbf{W}^\top \Big).
\end{equation}

Second, the \textbf{target distillation} loss distills knowledge from the frozen source adapter $G_{s}$. 
This ensures that the discriminative class structures and task-specific boundaries learned on the source domain are not erased during continuous adaptation:
\begin{equation}
  \mathcal{L}_{\text{pl}} = D_{\text{KL}}\! \big(G_{t}(\mathbf{z}) \cdot \mathbf{W}^\top \;\Big\|\; G_{s}(\mathbf{z}) \cdot \mathbf{W}^\top \Big)
\end{equation}

\noindent where $\mathbf{W}$ stacks the text embeddings containing the classes and prompt variations.
The total loss is defined as:
\begin{equation}
  \mathcal{L} = \mathcal{L}_{\text{zs}} + \lambda \, \mathcal{L}_{\text{pl}}
\end{equation}

\noindent where $\lambda$ weights the contribution of the target distillation term.

Adaptation is fully online: each arriving mini-batch of unlabeled target videos drives one or more optimization steps on $G_{t}$, and the updated target adapter immediately classifies that same batch. We do not reinitialize $G_{t}$ between batches; instead, it accumulates target-domain signals across the stream. During the evaluation of the batch, predictions are generated strictly using $G_{t}$ after the updates are applied.

\subsection{Intuition}
\label{sec:intutition}

Figure~\ref{fig:tadd-overview} summarizes the complete TADD pipeline. 
As shown in (a), the process begins with the offline source-domain pre-training of the projection adapter $G_{s}$, keeping the CLIP visual encoder frozen. During deployment, as depicted in (b), the online TTA pipeline is activated: unlabeled target batches drive updates to $G_{t}$ through the dual distillation losses, while the backbone and $G_{s}$ stay fixed.

Although these two distillation losses are straightforward, they play vital, complementary roles. 
The zero-shot distillation regularizes the network against forgetting CLIP's global vision-language semantics, which serve as a stable reference under severe domain shifts. 
Concurrently, the target distillation transfers the class-specific structure learned by $G_{s}$ on the source domain. 
This prevents the adaptation process from collapsing into solutions that are well-aligned with text but poorly separated in the visual feature space (and vice versa). 
Together, these losses steer $G_{t}$ toward representations that remain semantically grounded in CLIP's generalization capabilities while precisely matching the distinct characteristics of the target video distribution.

\section{Experiments and Results}
\label{sec:experiments}

In this section, we comprehensively evaluate our TADD framework.
Section~\ref{sec:experiments_setup} details our experimental setup.
Section~\ref{sec:results_cs} presents the main results obtained for standard closed-set test-time adaptation.
Section~\ref{sec:results_es} explores the extended settings of partial-set and sequential multi-target adaptation, demonstrating the robustness of our approach.
Finally, Section~\ref{sec:ablation} provides extensive ablation studies to validate our architectural choices, analyzing the impact of our dual distillation loss components, batch size variations, computational efficiency, and hyperparameter sensitivity.

\subsection{Experimental Setup}
\label{sec:experiments_setup}

This section describes the experimental protocol adopted to evaluate our TADD framework.
First, we outline the three video action recognition benchmarks utilized in our study: UCF-HMDB, Daily-DA, and Sports-DA.
Then, we introduce the baseline approaches selected for comparison. 
Lastly, we specify our implementation details, including the configuration of the frozen CLIP backbone and hyperparameters.

\paragraph{\textbf{Datasets.}}
We first report results on \textbf{UCF-HMDB\textsubscript{full}}~\cite{ucfhmdb}, which contains 3,205 videos spanning 12 action categories derived from HMDB51~\cite{hmdb51} and UCF101~\cite{ucf101}. Additionally, we consider \textbf{UCF-HMDB\textsubscript{partial}}~\cite{xu2021partial}, where the source domain contains 14 classes and the target domain contains 7 classes from HMDB51~\cite{hmdb51} and UCF101~\cite{ucf101}, following the official training and validation splits.
We also evaluate on \textbf{Daily-DA}~\cite{sportsda}, which comprises 18,949 videos from 8 daily action classes, constructed from four action recognition datasets: HMDB51~\cite{hmdb51}, ARID~\cite{arid}, MIT~\cite{mit}, and Kinetics~\cite{kinetics}.
Finally, we evaluate on \textbf{Sports-DA}~\cite{sportsda}, which aggregates three datasets, UCF101~\cite{ucf101}, Sports-1M~\cite{sports1m}, and Kinetics~\cite{kinetics} into a benchmark of 40,715 videos across 23 sports-related classes.
For brevity in our tables and subsequent text, we denote the domains by their initial letters: U (UCF101), H (HMDB51), A (ARID), M (MIT), K (Kinetics), and S (Sports-1M).

\paragraph{\textbf{Baselines.}}
Initially, we compare our methods against a \textit{Lower Bound}, that denotes the source-trained adapter evaluated on the target domain without any adaptation, and the \textit{Zero-shot}, that denotes the frozen CLIP model without any adapter.
Then, we compare our method against five representative TTA approaches spanning both image-based and video-specific settings, including TENT~\cite{wang2020tent}, SHOT~(online)~\cite{liang2020we}, ST2ST~\cite{fahim2024st2st}, ViTTA~\cite{lin2023video}, and BATCLIP~\cite{maharana2025batclip}. 
Originally a SFDA, SHOT~(online) is adapted for TTA by minimizing entropy and information maximization regularization at test-time, following Lin~et~al.~\cite{lin2023video}.
TENT, SHOT and BATCLIP are originally designed for image-based adaptation, while ST2ST and ViTTA are tailored for video action recognition. BATCLIP is the closest to our approach, as it adapts CLIP at test time.
Finally, we also compare against three SFDA methods: ATCoN~\cite{xu2022atcon}, EXTERN~\cite{xu2024extern}, and DALL-V~\cite{zara2023unreasonable}. These methods operate under a less restrictive setting, having access to the entire unlabeled target dataset for offline adaptation, whereas TTA must adapt online using only the current test batch. 
We adopted the same hyperparameters and training protocols as the original works. For the image-based TTA methods, we applied the strategy described in Section~\ref{sec:background} to obtain video-level predictions.

\paragraph{\textbf{Implementation details.}} 
We build upon the CLIP ViT-B/32 model with pre-trained weights from OpenAI. For each video, we uniformly sample 16 frames and resize to $224 \times 224$ pixels, forming a single clip per video. Frame-level features are extracted independently using the frozen CLIP visual encoder and then temporally aggregated via mean pooling to obtain a video-level representation. 
The source adapter $G_{s}$ is pre-trained on the source domain for 30 epochs. 
At test time, the CLIP backbone remains frozen and only the target adapter $G_{t}$ is updated. We use SGD with a fixed learning rate of $10^{-4}$ and perform $t$ gradient steps per batch. The target distillation weight $\lambda$ is set to 0.1, and the batch size is fixed at 8 across all experiments. 
All experiments are conducted on a single machine with an AMD Ryzen Threadripper PRO 5975WX 32-core CPU, 256 GB of DDR4 memory, and four NVIDIA RTX 5000 Ada Generation GPUs. The system runs Ubuntu 22.04.4 LTS with Linux kernel 6.8.0 and the ext4 file system.

\subsection{Closed-set Test-Time Adaptation}
\label{sec:results_cs}

In this section, we present closed-set TTA results, reporting top-1 accuracy across all source$\rightarrow$target directions.

\paragraph{\textbf{Results on HMDB51$\leftrightarrow$UCF101.}}
As shown in Table~\ref{tab:results}, the frozen CLIP model (zero-shot) establishes a strong baseline of 85.2\% on average, surpassing all standard TTA baselines. Our TADD achieves the best TTA average at 87.4\%, outperforming the strongest TTA baseline (BATCLIP) and zero-shot by +2.2 points. The improvement is most pronounced on H$\rightarrow$U, where TADD obtain 91.6\%—more than 9 points above BATCLIP and more than 4 points above zero-shot. On U$\rightarrow$H, BATCLIP attains the best TTA score, slightly above zero-shot, while TADD remains competitive at 83.2\%. Notably, image-based TTA methods improve only marginally over the lower bound, and ST2ST and ViTTA stay below the zero-shot average, suggesting that video TTA without a strong vision-language anchor fails to fully exploit this benchmark.

We also compare against 
SFDA methods.
Even with the frozen CLIP model (zero-shot) at 85.2\% average, offline SFDA still reaches higher accuracy; under the stricter online constraint, our TADD surpasses ATCoN by +4.9 points and narrows the gap to stronger SFDA approaches, like EXTERN and DALL-V. This highlights the effectiveness of our TADD framework, achieving competitive performance with SFDA methods while operating under significantly stricter adaptation conditions.

\begin{table}[!t]
  \centering
  \caption{Domain adaptation results on closed-set HMDB51$\leftrightarrow$UCF101.}
  \label{tab:results}
  \resizebox{0.75\linewidth}{!}{
  \begin{tabular}{c l c c c}
  \toprule
   & \textbf{Method} & \textbf{H$\rightarrow$U} & \textbf{U$\rightarrow$H} & \textbf{Avg.} \\
  \midrule

   & Lower Bound   & 85.6 & 73.9 & 79.8 \\
   & Zero-shot     & 87.5 & 82.8 & 85.2 \\
  \midrule

  \multirow{3}{*}{\rotatebox{90}{\footnotesize SFDA}}
   & ATCoN~\cite{xu2022atcon}     & 85.3 & 79.7 & 82.5  \\
   & EXTERN~\cite{xu2024extern}    & 91.9 & 88.9 & 90.4 \\
   & DALL-V~\cite{zara2023unreasonable}    & 93.1 & 88.9 & 91.0 \\
  \midrule

  \multirow{6}{*}{\rotatebox{90}{\footnotesize TTA}} 
   & TENT~\cite{wang2020tent}      & 78.1 & 78.3 & 78.2 \\
   & SHOT~\cite{liang2020we}      & 77.8 & 79.4 & 78.6 \\
   & ST2ST~\cite{fahim2024st2st}     & 80.9 & 78.6 & 79.8 \\
   & ViTTA~\cite{lin2023video}     & 82.0 & 78.7 & 80.3 \\
   & BATCLIP~\cite{maharana2025batclip}   & 82.0 & \textbf{85.3} & 83.6 \\
   & TADD (ours)      & \textbf{91.6} & 83.2 & \textbf{87.4} \\

  \bottomrule
  \end{tabular}
  }
  \end{table}

\begin{table*}[t]
\centering
\caption{Domain adaptation results on closed-set Daily-DA.}
\label{tab:multi_domain}
\resizebox{0.9\textwidth}{!}{
\begin{tabular}{clccccccccccccc}
\toprule
 & \textbf{Method} & \textbf{H$\rightarrow$K} & \textbf{H$\rightarrow$M} & \textbf{H$\rightarrow$A} & 
\textbf{A$\rightarrow$M} & \textbf{A$\rightarrow$K} & \textbf{A$\rightarrow$H} & 
\textbf{M$\rightarrow$K} & \textbf{M$\rightarrow$A} & \textbf{M$\rightarrow$H} & 
\textbf{K$\rightarrow$A} & \textbf{K$\rightarrow$H} & \textbf{K$\rightarrow$M} & \textbf{Avg.} \\
\midrule

 & Lower Bound  & 40.0 & 36.8 & 20.5 & 26.5 & 31.3 & 35.4 & 67.1 & 25.9 & 45.0 & 26.1 & 42.9 & 37.0 & 36.1 \\
 & Zero-shot    & 63.1 & 37.5 & 25.2 & 37.5 & 63.1 & 39.2 & 63.1 & 25.2 & 39.2 & 25.2 & 39.2 & 37.5 & 38.3 \\
\midrule

\multirow{3}{*}{\rotatebox{90}{\footnotesize SFDA}}
 & ATCoN~\cite{xu2022atcon}     & 48.5 & 30.7 & 17.9 & 17.2 & 31.0 & 26.7 & 57.7 & 27.2 & 47.3 & 17.2 & 48.2 & 32.5 & 33.5 \\
 & EXTERN~\cite{xu2024extern}    & 57.6 & 40.7 & 26.2 & 18.2 & 51.4 & 26.2 & 68.1 & 18.1 & 53.7 & 23.9 & 55.8 & 35.2 & 39.6 \\
 & DALL-V~\cite{zara2023unreasonable}    & 76.7 & 47.0 & 24.0 & 45.7 & 75.0 & 57.9 & 78.1 & 24.0 & 65.4 & 24.0 & 52.5 & 47.0 & 51.4 \\
\midrule

\multirow{6}{*}{\rotatebox{90}{\footnotesize TTA}} 
 & TENT~\cite{wang2020tent}         & 40.7 & 36.5 & 27.7 & 25.5 & 33.4 & 40.4 & \textbf{71.1} & 25.9 & 57.1 & 25.6 & 41.7 & 39.3 & 37.9 \\
 & SHOT~\cite{liang2020we}         & 42.9 & 38.0 & 24.5 & 25.8 & 29.0 & 42.5 & 71.0 & 29.3 & 59.6 & 22.7 & 42.9 & 39.0 & 38.5 \\
 & ST2ST~\cite{fahim2024st2st}        & 40.8 & 36.8 & 27.9 & 25.0 & 31.4 & 43.3 & 69.9 & 30.0 & 59.6 & 25.5 & 40.8 & 41.3 & 38.8 \\
 & ViTTA~\cite{lin2023video}        & 42.2 & 37.3 & 26.5 & 25.3 & 32.8 & 41.3 & 70.8 & 27.5 & 57.5 & 25.1 & 42.1 & 38.8 & 38.0 \\
 & BATCLIP~\cite{maharana2025batclip}      & 65.9 & 39.5 & 24.3 & 39.5 & \textbf{65.9} & \textbf{47.9} & 65.9 & 24.3 & 47.9 & 24.3 & \textbf{47.9} & 39.5 & 43.7 \\
 & TADD (ours)         & \textbf{69.3} & \textbf{46.4} & \textbf{31.1} & \textbf{44.3} & 63.6 & 46.3 & 69.0 & \textbf{31.3} & \textbf{47.8} & \textbf{31.6} & 47.4 & \textbf{44.9} & \textbf{46.3} \\

\bottomrule
\end{tabular}
}
\end{table*}

\paragraph{\textbf{Results on Daily-DA.}}
Table~\ref{tab:multi_domain} reports results across all 12 source$\rightarrow$target directions on Daily-DA, the most challenging benchmark due to the severe illumination gaps introduced by the ARID domain. Our TADD achieves the highest average accuracy among TTA methods at 46.3\%, outperforming BATCLIP by +2.6 points and the best non-CLIP baseline (ST2ST) by +7.5 points. 
Our method achieves the best accuracy in 8 out of 12 directions, while BATCLIP leads on 3 directions (A$\rightarrow$K, A$\rightarrow$H, K$\rightarrow$H). Image-based TTA methods (TENT, SHOT) and even ViTTA and ST2ST provide only marginal improvements over the lower bound and zero-shot baselines on this benchmark, highlighting the difficulty of adapting under extreme domain shifts without leveraging CLIP's vision-language alignment.
Compared to the offline SFDA baselines, TADD remarkably surpasses both ATCoN (+12.8 points) and EXTERN (+6.7 points), despite operating in the strictly online TTA setting.

\paragraph{\textbf{Results on Sports-DA.}}
Table~\ref{tab:s1m_results} details results on the Sports-DA benchmark. A distinctive feature here is that CLIP's zero-shot performance substantially exceeds the source-trained lower bound, indicating strong representation of sports classes in CLIP's pre-training data. 
Consequently, conventional TTA methods that do not leverage knowledge of the CLIP model degrade performance relative to the zero-shot baseline. 
In contrast, our dual distillation approach explicitly anchors the source adapter to CLIP's zero-shot predictions, allowing it to benefit from this strong prior while incorporating source-domain knowledge. Our TADD achieves the best overall average accuracy of 88.4\%, improving over BATCLIP by +3.0 points and over the zero-shot baseline by +1.3 points.
Notably, on this benchmark, our strictly online TTA approach (TADD) surpasses all offline SFDA baselines (ATCoN, EXTERN, DALL-V).

\begin{table}[!t]
\centering
\caption{Domain adaptation results on closed-set Sports-DA.}
\label{tab:s1m_results}
\resizebox{\linewidth}{!}{
\begin{tabular}{clccccccc}
\toprule
 & \textbf{Method} & \textbf{S$\rightarrow$K} & \textbf{S$\rightarrow$U} & \textbf{U$\rightarrow$S} & \textbf{U$\rightarrow$K} & \textbf{K$\rightarrow$S} & \textbf{K$\rightarrow$U} & \textbf{Avg.} \\
\midrule

 & Lower Bound  & 66.6 & 80.1 & 68.0 & 73.6 & 73.5 & 72.2 & 72.8 \\
 & Zero-shot    & 87.1 & 89.1 & 79.5 & 87.1 & 79.5 & 89.1 & 87.1 \\
\midrule

\multirow{3}{*}{\rotatebox{90}{\footnotesize SFDA}}
 & ATCoN~\cite{xu2022atcon}     & 76.0 & 90.6 & 47.9 & 65.2 & 69.7 & 93.6 & 73.8 \\
 & EXTERN~\cite{xu2024extern}    & 82.2 & 95.4 & 72.7 & 81.2 & 73.8 & 93.7 & 83.2 \\
 & DALL-V~\cite{zara2023unreasonable}    & 82.3 & 88.8 & 75.9 & 81.2 & 77.7 & 88.0 & 82.3 \\
\midrule

\multirow{6}{*}{\rotatebox{90}{\footnotesize TTA}} 
 & TENT~\cite{wang2020tent}         & 82.0 & 89.1 & 72.7 & 76.7 & 55.6 & 70.3 & 74.7 \\
 & SHOT~\cite{liang2020we}         & 81.1 & 88.8 & 73.6 & 76.9 & 63.2 & 71.3 & 75.2 \\
 & ST2ST~\cite{fahim2024st2st}        & 80.2 & 92.6 & 71.0 & 74.3 & 61.9 & 66.2 & 72.7 \\
 & ViTTA~\cite{lin2023video}        & 80.6 & 89.7 & 71.8 & 75.9 & 62.7 & 72.0 & 74.0 \\
 & BATCLIP~\cite{maharana2025batclip}      & 85.4 & \textbf{91.5} & 80.0 & 85.4 & 80.0 & \textbf{91.5} & 85.4 \\
 & TADD (ours)         & \textbf{88.5} & 91.0 & \textbf{80.5} & \textbf{88.3} & \textbf{80.4} & 91.3 & \textbf{88.4} \\

\bottomrule
\end{tabular}
}
\end{table}

\subsection{Partial-set and Multi-target Adaptation}
\label{sec:results_es}

Beyond the standard closed-set scenario, we evaluate our TADD framework under two more challenging adaptation settings, partial-set and multi-target, where the source and target label spaces do not fully overlap. These settings test the robustness of TTA methods when faced with category mismatch between domains.

\paragraph{\textbf{Partial-set Adaptation.}} In this setting, the source domain contains 14 overlapping classes between UCF101 and HMDB51, while the target domain uses only a strict subset of 7 classes (the first 7 in alphabetical order), yielding two adaptation tasks: U-14$\rightarrow$H-7 and H-14$\rightarrow$U-7. This evaluates whether TTA methods can adapt to a target domain whose label space is a strict subset of the source domain.
As shown in Table~\ref{tab:ucf_hmdb}, the CLIP model (zero-shot) performs strongly here, since the 7 target classes are a clean subset that CLIP's text embeddings can discriminate well. Most conventional TTA methods (TENT, SHOT, ST2ST, ViTTA) fail to match zero-shot performance, with accuracies ranging from 69.3\% to 79.9\%, likely because adapting without awareness of the label space mismatch introduces noise from the irrelevant source classes. BATCLIP performs competitively at 89.3\%, leveraging CLIP's vision-language alignment.
Our TADD achieves the highest accuracy of 93.7\%, surpassing even the zero-shot baseline. This demonstrates that the dual distillation objective successfully combines the target's discriminative knowledge with CLIP's zero-shot prior: the zero-shot signal provides a clean class structure over the 7 target categories, while the target distillation contributes source-domain visual discrimination, yielding complementary gains.

\begin{table}[h!]
\centering
\caption{Domain adaptation results on partial-set UCF$\leftrightarrow$HMDB$_{partial}$.}
\label{tab:ucf_hmdb}
\resizebox{0.65\linewidth}{!}{
\begin{tabular}{lccc}
\toprule
\textbf{Method} & \textbf{U$\rightarrow$H} & \textbf{H$\rightarrow$U} & \textbf{Avg.} \\
\midrule
Lower Bound  & 54.8 & 86.1 & 70.4 \\
Zero-shot    & 88.1 & 94.5 & 91.3 \\
\midrule
TENT~\cite{wang2020tent}         & 69.1 & 87.2 & 78.1 \\
SHOT~\cite{liang2020we}         & 67.6 & 71.1 & 69.3 \\
ST2ST~\cite{fahim2024st2st}        & 70.0 & 89.9 & 79.9 \\
ViTTA~\cite{lin2023video}        & 69.1 & 88.6 & 78.8 \\
BATCLIP~\cite{maharana2025batclip}      & 83.8 & 94.7 & 89.3 \\
TADD (ours)         & \textbf{88.6} & \textbf{98.9} & \textbf{93.7} \\
\bottomrule
\end{tabular}
}
\end{table}

\begin{table*}[t]
\centering
\caption{Multi-target adaptation results on Daily-DA under different data orderings. We report mean $\pm$ std across orderings.}
\label{tab:order_effects}
\resizebox{0.8\textwidth}{!}{
\begin{tabular}{l cccc cccc cccc}
\toprule
 & \multicolumn{4}{c}{\textbf{A$\rightarrow$\{K,M,H\}}} & \multicolumn{4}{c}{\textbf{K$\rightarrow$\{A,M,H\}}} & \multicolumn{4}{c}{\textbf{H$\rightarrow$\{A,K,M\}}} \\
\cmidrule(lr){2-5} \cmidrule(lr){6-9} \cmidrule(lr){10-13}
\textbf{Method} & S.D. & S.C. & Rnd. & Avg. & S.D. & S.C. & Rnd. & Avg. & S.D. & S.C. & Rnd. & Avg. \\
\midrule

Lower Bound & \multicolumn{4}{c}{30.3} & \multicolumn{4}{c}{27.0} & \multicolumn{4}{c}{32.1} \\
Zero-shot   & \multicolumn{4}{c}{52.3} & \multicolumn{4}{c}{\textbf{34.9}} & \multicolumn{4}{c}{43.6} \\
\midrule

TENT~\cite{wang2020tent}    & 30.9 & 31.8 & 30.9 & 31.4{\scriptsize$\pm$0.6} & 28.7 & 26.3 & 29.9 & 29.8{\scriptsize$\pm$0.1} & 24.6 & 30.1 & 25.6 & 25.6{\scriptsize$\pm$2.9} \\
ST2ST~\cite{fahim2024st2st}   & 30.3 & 32.5 & 31.1 & 31.1{\scriptsize$\pm$1.1} & 30.1 & 31.9 & 20.9 & 30.1{\scriptsize$\pm$5.9} & 31.7 & 35.5 & 32.5 & 32.5{\scriptsize$\pm$2.0} \\
ViTTA~\cite{lin2023video}   & 30.0 & 29.7 & 29.8 & 29.8{\scriptsize$\pm$0.1} & 29.9 & 30.5 & 29.1 & 29.9{\scriptsize$\pm$0.7} & 32.2 & 33.1 & 33.8 & 33.1{\scriptsize$\pm$0.9} \\
BATCLIP~\cite{maharana2025batclip} & 55.6 & 55.4 & 55.4 & 55.4{\scriptsize$\pm$0.1} & 31.2 & 32.7 & 30.7 & 31.2{\scriptsize$\pm$1.0} & 40.5 & 40.0 & 39.5 & 40.0{\scriptsize$\pm$0.5} \\
TADD (ours)    & \textbf{56.5} & \textbf{58.7} & \textbf{56.3} & \textbf{56.5{\scriptsize$\pm$1.3}} & 32.6 & 34.6 & 32.1 & 32.6{\scriptsize$\pm$1.3} & \textbf{45.3} & \textbf{48.7} & \textbf{44.8} & \textbf{45.3{\scriptsize$\pm$2.1}} \\

\bottomrule
\end{tabular}
}
\end{table*}

\paragraph{\textbf{Multi-target Adaptation.}} We evaluate our TADD in a a sequential multi-target setting, where the model is exposed to multiple target domains continuously. Table~\ref{tab:order_effects} reports results on Daily-DA under three data orderings: (\textit{i}) \textbf{Sequential Domain} (S.D.), where domains are observed one at a time; (\textit{ii}) \textbf{Sequential Class} (S.C.), where samples are grouped by class across domains; and (\textit{iii}) \textbf{Random} (Rnd.), where samples are fully shuffled.
TADD achieves the best performance across most settings, outperforming BATCLIP on average. 
Also, TADD demonstrates stable performance across different orderings, with lower variance than prior methods.
In contrast, conventional TTA methods struggle under sequential domain shifts, often performing close to the lower bound. 
While BATCLIP remains competitive, its performance degrades under certain orderings. 
Our TADD framework mitigates this by leveraging both zero-shot and target distillation, improving robustness and generalization across multiple target domains.

\subsection{Ablation Studies}
\label{sec:ablation}

In this section, we conduct comprehensive ablation studies to validate the architectural choices and to evaluate the robustness of our TADD framework.
We begin analyzing the effects of our proposed loss components.
Next, we evaluate the stability of our method to variations of test-time batch sizes.
We then assess the computational efficiency of TADD.
Finally, we examine the sensitivity of our framework to key hyperparameters.

\paragraph{\textbf{Effect of Loss Components}} 
Table~\ref{tab:ablation_loss} isolates the impact of aligning adapted logits solely with CLIP's frozen zero-shot predictions ($\mathcal{L}_{zs}$) versus adding distillation from the source-trained adapter ($\mathcal{L}_{pl}$). The zero-shot-only setup ($\mathcal{L}_{zs}$) degrades performance on H$\rightarrow$A and H$\rightarrow$M relative to the frozen CLIP model (zero-shot), while providing only marginal gains on H$\rightarrow$K, indicating that global CLIP alignment alone is insufficient under strong domain shifts.
In contrast, combining both loss terms yields the best performance across all splits and matches the closed-set results for the same directions. Overall, while zero-shot distillation ($\mathcal{L}_{zs}$) provides a stable semantic anchor, the adapter-to-adapter distillation term ($\mathcal{L}_{pl}$) is crucial for preserving source-specific discrimination.

\begin{table}[h!]
\centering
\caption{Ablation studies of loss components on Daily-DA.}
\label{tab:ablation_loss}
\setlength{\tabcolsep}{6pt}
\begin{tabular}{lccc}
\toprule
\textbf{Method} & \textbf{H$\rightarrow$A} & \textbf{H$\rightarrow$K} & \textbf{H$\rightarrow$M} \\
\midrule
Zero-shot & 30.2 & 67.3 & 43.8 \\
+ $\mathcal{L}_{zs}$ & 25.4 & 68.4 & 41.8 \\
+ $\mathcal{L}_{pl}$ & \textbf{31.1} & \textbf{69.3} & \textbf{46.4} \\
\bottomrule
\end{tabular}
\end{table}

\paragraph{\textbf{Impact of Batch Size}} 
Because TTA relies on whatever unlabeled batch arrives at inference, stability across batch sizes is critical.
Table~\ref{tab:batch_size} reports closed-set accuracy on HMDB$\rightarrow$UCF for batch sizes ranging from 1 to 16.
Our TADD improves steadily as the batch grows, peaking at a batch size of 16. 
This suggests that larger batches provide more stable gradient estimates for the adapter updates. 
BATCLIP peaks early and then loses ground as the batch increases, while ST2ST collapses with a single sample per step and only becomes competitive once the batch is at least two. TENT stays nearly flat, whereas ViTTA drifts down modestly as the batch grows. 
TADD remains highly robust even at a batch size of 1 and scales favorably as more target samples become available.

\begin{table}[h!]
\centering
\caption{Impact of batch size on HMDB$\rightarrow$UCF.}
\label{tab:batch_size}
\setlength{\tabcolsep}{5pt}
\begin{tabular}{lcccccc}
\toprule
& \multicolumn{5}{c}{\textbf{Batch Size}} & \\
\cmidrule(lr){2-6}
\textbf{Method} & \textbf{1} & \textbf{2} & \textbf{4} & \textbf{8} & \textbf{16} & \textbf{Avg.} \\
\midrule
TENT~\cite{wang2020tent}    & 80.4 & 80.4 & 80.6 & 80.1 & 81.5 & 80.6 \\
ST2ST~\cite{fahim2024st2st}   & 17.2 & 79.7 & 80.9 & 80.9 & 81.3 & 68.0 \\
ViTTA~\cite{lin2023video}   & 83.2 & 82.8 & 82.1 & 82.0 & 81.6 & 82.3 \\
BATCLIP~\cite{maharana2025batclip} & 66.4 & 89.0 & 86.7 & 83.7 & 84.4 & 82.0 \\
TADD (ours)    & \textbf{88.6} & \textbf{89.0} & \textbf{90.7} & \textbf{91.6} & \textbf{91.7} & \textbf{90.3} \\
\bottomrule
\end{tabular}
\end{table}

\paragraph{\textbf{Computational Efficiency}} 
As shown in Table~\ref{tab:efficiency}, our TADD is highly efficient in terms of adaptation time per batch. Although TADD contains more trainable parameters than BATCLIP (131K vs 65K), it achieves lower runtime (0.143s vs 0.179s). This is because TADD updates only a lightweight adapter on top of frozen visual features, whereas BATCLIP requires gradient propagation through the entire CLIP backbone to update its LayerNorm parameters, resulting in a higher per-iteration computational cost.

\begin{table}[h!]
\centering
\caption{Computational complexity of TTA models}
\label{tab:efficiency}
\resizebox{0.4\textwidth}{!}{
\begin{tabular}{lcc}
\toprule
\textbf{Method} & \textbf{Time/Batch (s)} & \textbf{\#Train. Params} \\
\midrule
SHOT~\cite{liang2020we}            & 0.189 & 151M \\
ST2ST~\cite{fahim2024st2st}           & 0.297 & 151M \\
ViTTA~\cite{lin2023video}           & 0.282 & 151M \\
BATCLIP~\cite{maharana2025batclip}         & 0.179 & 65K \\
TADD (ours) (iter=20)  & \textbf{0.143} & 131K \\
\bottomrule
\end{tabular}
}
\end{table}

\begin{figure}[h!]
  \centering
  \begin{subfigure}[t]{\columnwidth}
    \centering
    \includegraphics[width=\linewidth,height=0.2\textheight,keepaspectratio]{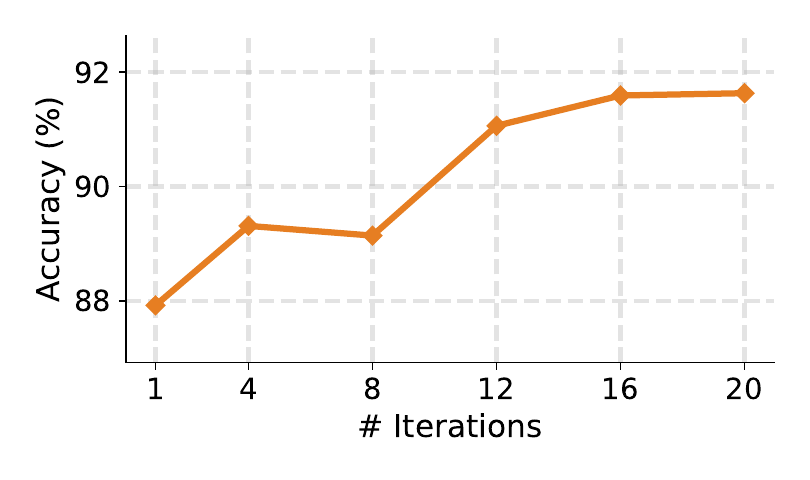}
    \caption{Accuracy as a function of the number of iterations}
    \label{fig:iter_hyperparam_a}
  \end{subfigure}
  \\
  \begin{subfigure}[t]{\columnwidth}
    \centering
    \includegraphics[width=\linewidth,height=0.2\textheight,keepaspectratio]{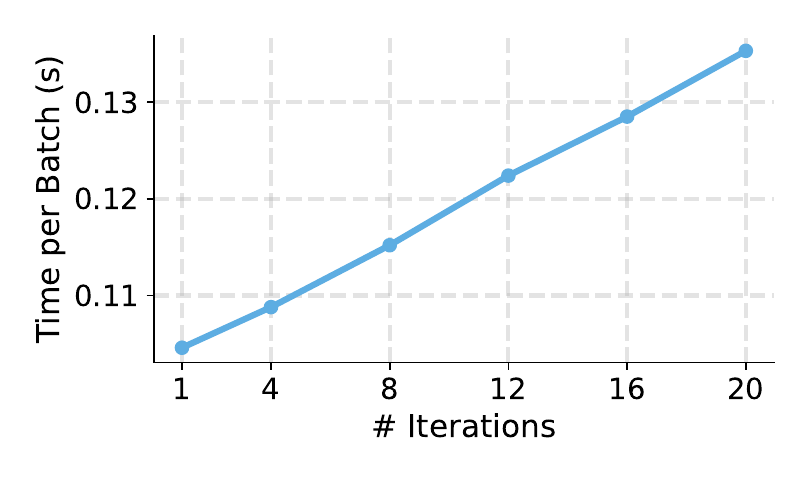}
    \caption{Time per batch as a function of the number of iterations.}
    \label{fig:iter_hyperparam_b}
  \end{subfigure}
  \caption{Sensitivity to the per-batch adaptation iteration count on HMDB$\rightarrow$UCF.}
  \label{fig:iter_hyperparam}
\end{figure}

\paragraph{\textbf{Sensitivity to Hyperparameter}} 
The number of gradient steps applied to each target batch is a key hyperparameter, where more iterations can improve adaptation quality but also increase latency. Figure~\ref{fig:iter_hyperparam} reports, on HMDB$\rightarrow$UCF, how accuracy varies with this iteration budget (a) and how the average time per batch scales with the same setting (b). 
While our TADD achieves higher accuracy with more iterations, it remains faster than state-of-the-art methods even at higher iteration counts due to its lightweight architectural design.

\section{Conclusions}
\label{sec:conclusions}

Adapting video action recognition models at test time is notoriously challenging, especially when target video streams exhibit complex temporal dynamics and severe domain shifts. 
In spite of all the advances, existing TTA methods often transfer poorly to these restrictive settings.
To address this critical gap, we introduced \textbf{Test-time Adaptation via Dual Distillation (TADD)}, a strictly online, parameter-efficient TTA framework. By leveraging a frozen CLIP backbone and updating only a lightweight projection adapter during inference, TADD effectively circumvents catastrophic forgetting and computational bottlenecks. This stability is driven by a novel dual distillation objective: \textit{zero-shot distillation} acts as a semantic anchor to preserve CLIP's domain-agnostic vision-language alignment, while \textit{target distillation} serves as a discriminative anchor to retain the task-specific boundaries learned from the source domain.

Empirical evaluations across three diverse benchmarks—UCF-HMDB, Daily-DA, and Sports-DA—demonstrate that our TADD consistently sets a new state-of-the-art for video TTA. TADD not only outperforms standard image-based TTA baselines but also rivals or surpasses complex SFDA methods, despite the latter operating under much less restrictive offline settings. Furthermore, our TADD framework exhibits exceptional robustness in restrictive and complex scenarios, including partial-set and sequential multi-target adaptation, maintaining stable accuracy across varying batch sizes and streaming orderings.

For future work, we plan to extend the principles of TADD to the paradigm of universal TTA, where the exact nature of the target distribution shift (e.g., concurrent covariate, label, and open-set shifts) is completely unknown prior to deployment. Additionally, exploring tighter integration with explicit temporal modeling architectures—moving beyond simple temporal frame pooling—presents a promising avenue to further enhance online video TTA.

\begin{acks}
This research was supported by São Paulo Research Foundation - FAPESP (grants 2023/17577-0, 2024/04500-2, and 2024/22985-3) and National Council for Scientific and Technological Development - CNPq (grants 315220/2023-6, 420442/2023-5, and 444982/2024-8).
\end{acks}



\appendix

\section{Sensitivity to Data Ordering in Test-Time Adaptation}

In real-world deployments, unlabeled target data rarely arrives in a purely random order. Video streams often exhibit strong temporal or categorical correlation. To rigorously evaluate the robustness of our method against such non-i.i.d. data streams, we repeat the closed-set adaptation experiments while varying only how mini-batches are formed from the target stream.

Tables~\ref{tab:ordering_hu}~and~\ref{tab:ordering_uh} report the top-1 accuracy under four distinct batch construction strategies, alongside their overall average. The streaming protocols are defined as follows:
\begin{enumerate}
    \item \textbf{Random (default):} Target examples are drawn uniformly at random into each batch.
    \item \textbf{Sequential (linear):} Examples are visited in a fixed class order (e.g., all of class 0, then all of class 1).
    \item \textbf{Sequential Group (double linear):} Each class is split into two halves, $A$ and $B$. All $A$ segments are streamed across all classes before any $B$ segments are introduced.
    \item \textbf{Batch-Class:} Each batch contains instances from a single class, with the order of classes randomized across the stream.
\end{enumerate}
For reference, we also report the results for the Zero-shot (frozen CLIP) baseline. As it performs no adaptation at test time, its accuracy remains completely invariant to the batch construction strategy.

\begin{table*}[!htb]
  \centering
  \caption{Sensitivity to data ordering on HMDB51$\rightarrow$UCF101 (closed-set). Accuracy (\%) for each batch construction strategy.}
  \label{tab:ordering_hu}
  \setlength{\tabcolsep}{5pt}
  \begin{tabular}{lccccc}
    \toprule
    \textbf{Method} & \textbf{Random} & \textbf{Sequential} & \textbf{Sequen. Group} & \textbf{Batch-Class} & \textbf{Avg.} \\
    \midrule
    Zero-shot   & \multicolumn{5}{c}{89.5} \\
    \midrule
    TENT~\cite{wang2020tent}           & 78.1 & 80.9 & 81.8 & 81.1 & 81.0 \\
    ST2ST~\cite{fahim2024st2st}        & 80.9 & 83.7 & 79.3 & 83.4 & 82.1 \\
    ViTTA~\cite{lin2023video}          & 82.0 & 81.3 & 81.8 & 81.3 & 81.5 \\
    BATCLIP~\cite{maharana2025batclip} & 82.0 & 85.1 & 88.8 & 85.8 & 85.5 \\
    TADD (ours)                        & \textbf{91.6} & \textbf{93.7} & \textbf{93.3} & \textbf{93.2} & \textbf{93.3} \\
    \bottomrule
  \end{tabular}%
\end{table*}

\begin{table*}[!htb]
  \centering
  \caption{Sensitivity to data ordering on UCF101$\rightarrow$HMDB51 (closed-set). Accuracy (\%) for each batch construction strategy.}
  \label{tab:ordering_uh}
  \setlength{\tabcolsep}{5pt}
  \begin{tabular}{lccccc}
    \toprule
    \textbf{Method} & \textbf{Random} & \textbf{Sequential} & \textbf{Sequen. Group} & \textbf{Batch-Class} & \textbf{Avg.} \\
    \midrule
    Zero-shot   & \multicolumn{5}{c}{82.8} \\
    \midrule
    TENT~\cite{wang2020tent}           & 78.3 & 78.1 & 79.8 & 78.1 & 78.2 \\
    ST2ST~\cite{fahim2024st2st}        & 78.6 & 80.0 & 76.1 & 77.7 & 78.2 \\
    ViTTA~\cite{lin2023video}          & 78.7 & 78.1 & 78.3 & 78.1 & 78.2 \\
    BATCLIP~\cite{maharana2025batclip} & \textbf{85.3} & 85.1 & 82.5 & 82.2 & 83.8 \\
    TADD (ours)                        & 83.2 & \textbf{90.3} & \textbf{88.7} & \textbf{90.5} & \textbf{89.5} \\
    \bottomrule
  \end{tabular}%
\end{table*}

As shown in Table~\ref{tab:ordering_hu}, TADD achieves the highest accuracy on HMDB51$\rightarrow$UCF101 under every batch ordering. Notably, there is only a narrow performance spread between the strictly random setting and the highly structured streams. By contrast, prior TTA baselines fall well below zero-shot performance on average, whereas TADD surpasses the zero-shot baseline by several points.

For the UCF101$\rightarrow$HMDB51 benchmark, the results in  Table~\ref{tab:ordering_uh} show a more asymmetric picture. While BATCLIP attains the highest accuracy on strictly random batches (slightly ahead of our method), TADD achieves vastly superior results under the linear, double-linear, and single-class-per-batch streams. Averaged across all four orderings, our approach leads by a substantial margin. This suggests that TADD's overall gains stem from its exceptional stability when the stream exhibits strong class structure—a setting frequently encountered when video frames or clips are organized temporally. In summary, our approach combines competitive random-batch performance with markedly more robust behavior on highly structured streams.

\section{Additional Visualizations}

\begin{figure*}[!htb]
  \centering
  \captionsetup[subfigure]{font=footnotesize, skip=4pt, belowskip=2pt}
  \footnotesize
  \begin{subfigure}[t]{0.22\textwidth}
    \centering
    \includegraphics[width=\linewidth]{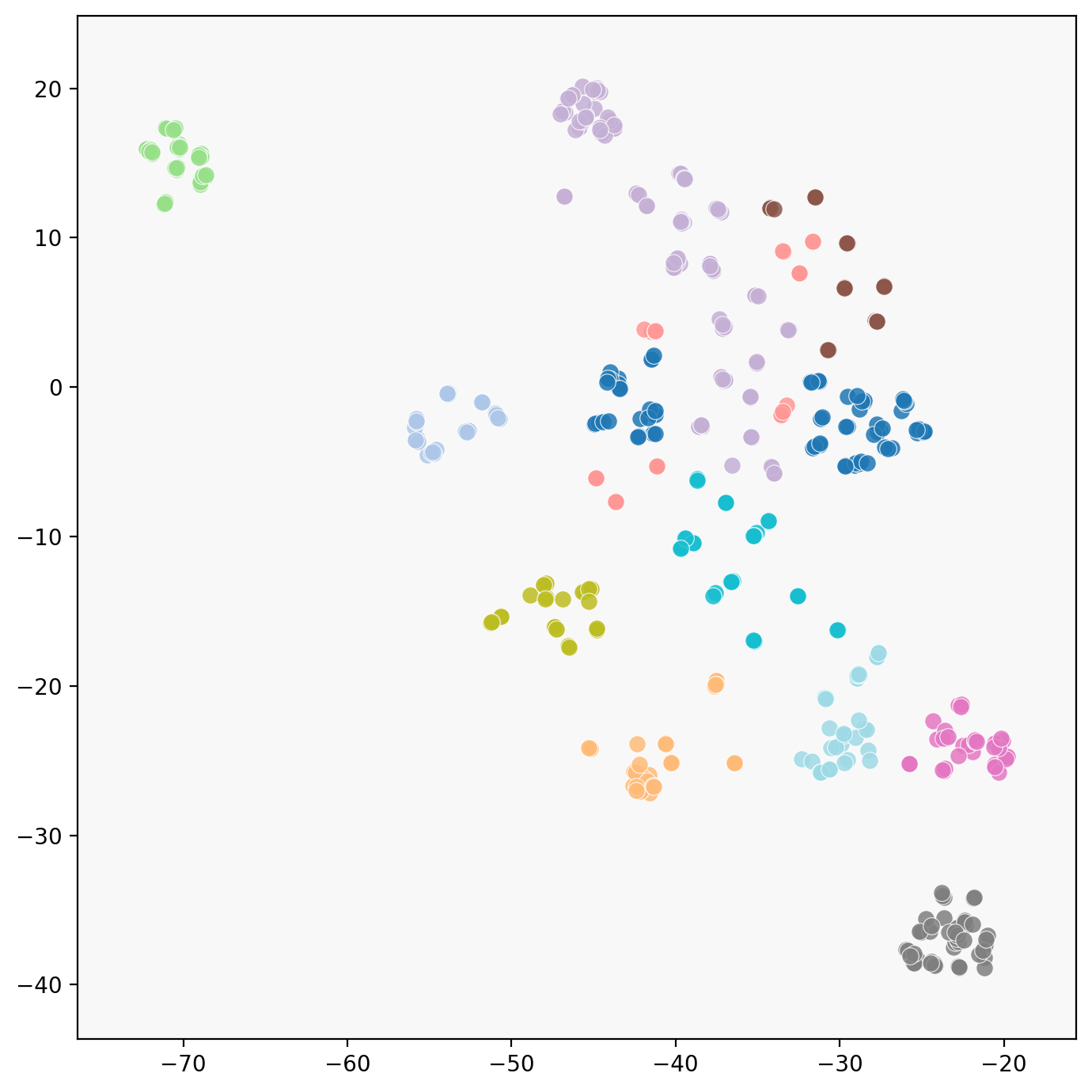}
    \caption{Zero-shot (CLIP~\cite{radford2021learning})}
    \label{fig:appendix_tsne_a}
  \end{subfigure}\hfill
  \begin{subfigure}[t]{0.22\textwidth}
    \centering
    \includegraphics[width=\linewidth]{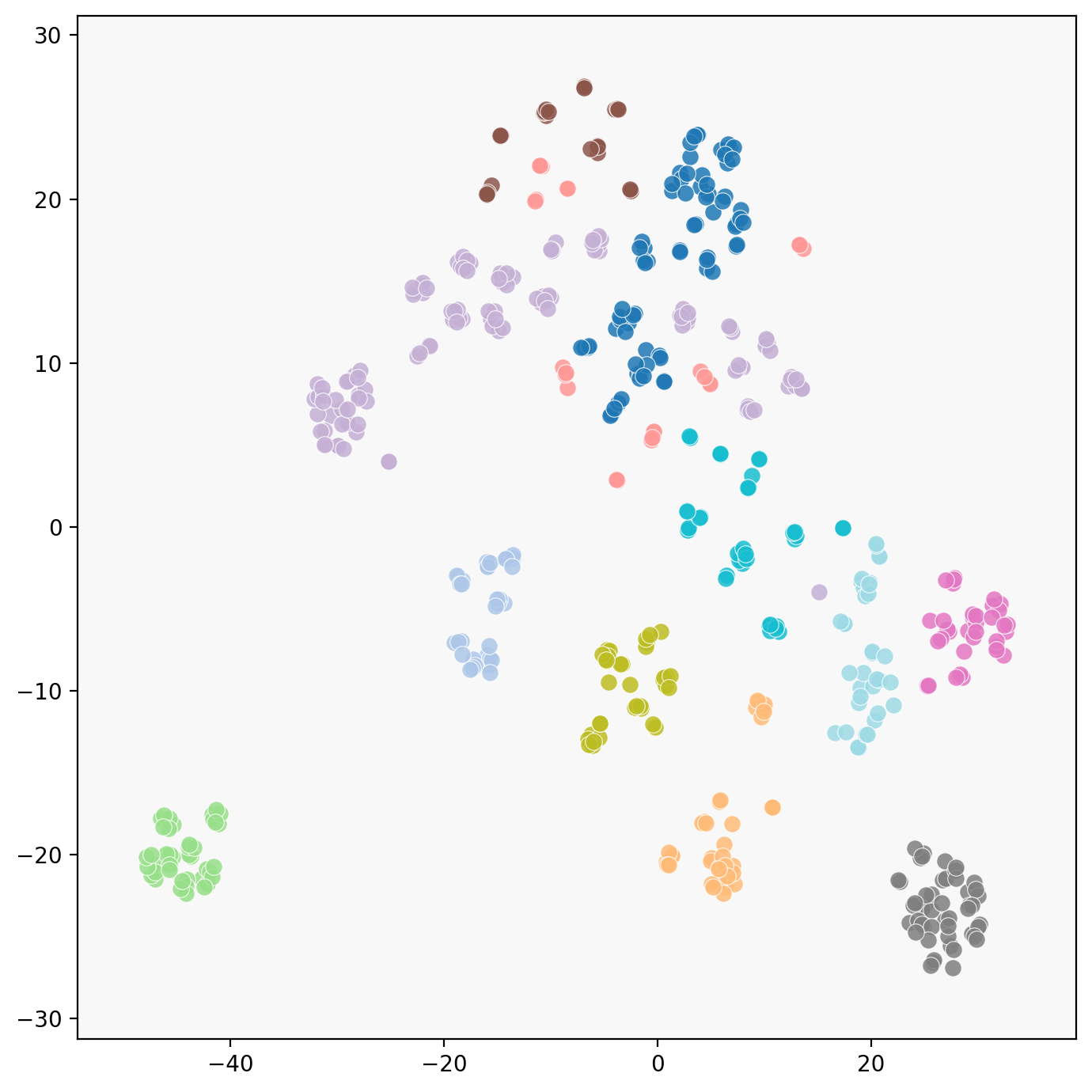}
    \caption{ViTTA~\cite{lin2023video}}
    \label{fig:appendix_tsne_b}
  \end{subfigure}\hfill
  \begin{subfigure}[t]{0.22\textwidth}
    \centering
    \includegraphics[width=\linewidth]{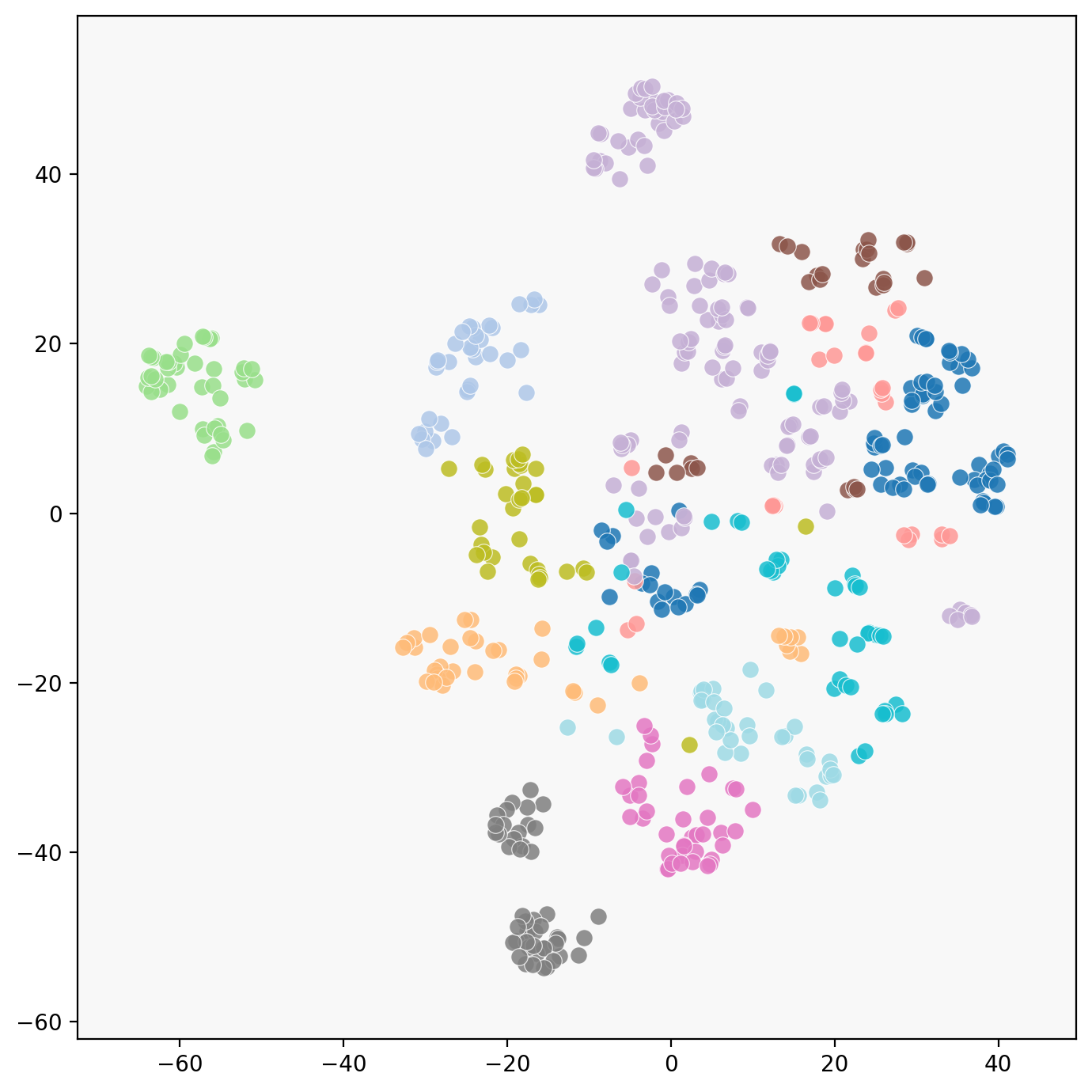}
    \caption{BATCLIP~\cite{maharana2025batclip}}
    \label{fig:appendix_tsne_c}
  \end{subfigure}\hfill
  \begin{subfigure}[t]{0.22\textwidth}
    \centering
    \includegraphics[width=\linewidth]{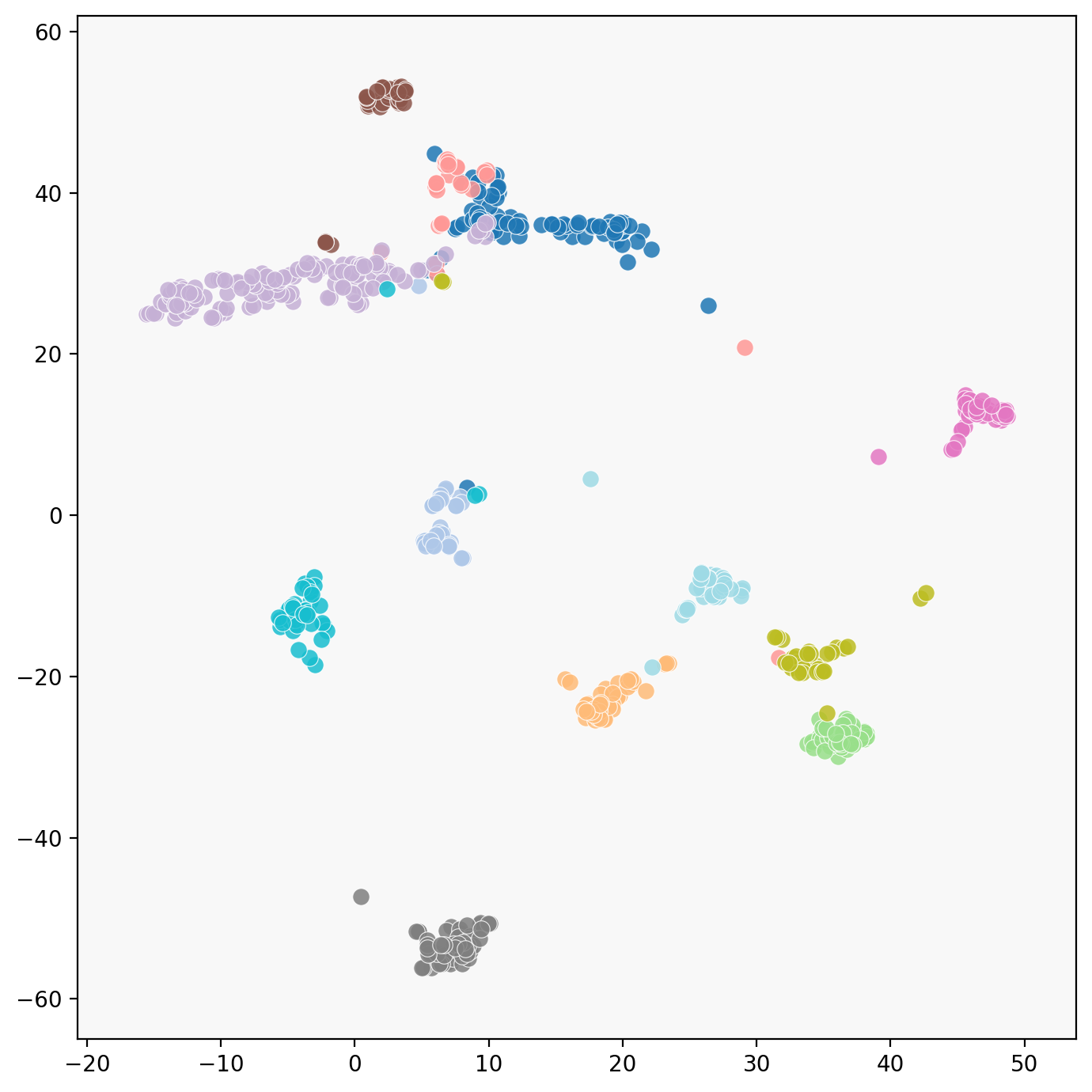}
    \caption{TADD (ours)}
    \label{fig:appendix_tsne_d}
  \end{subfigure}

  \par\bigskip

  \begin{subfigure}[t]{0.22\textwidth}
    \centering
    \includegraphics[width=\linewidth]{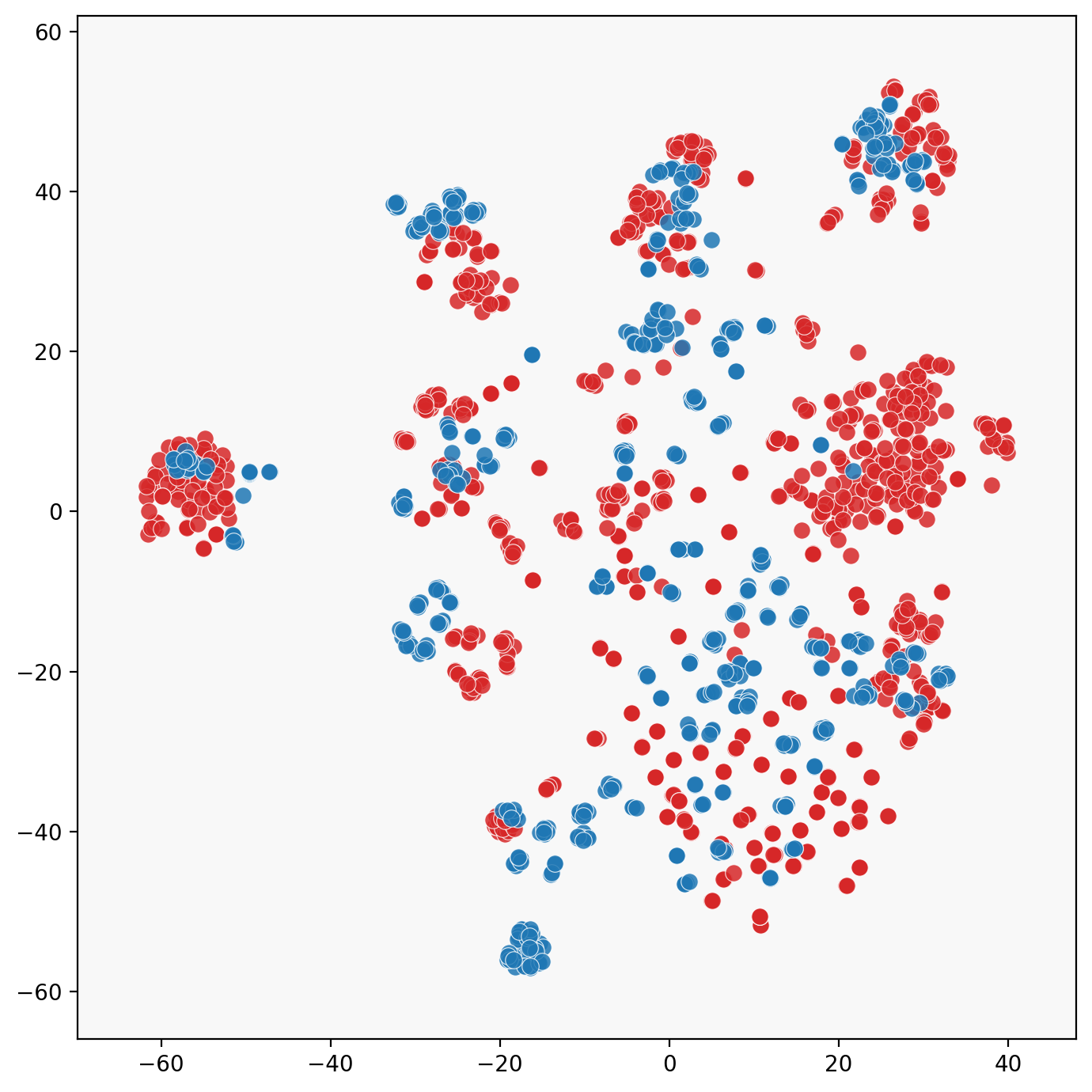}
    \caption{Zero-shot (CLIP~\cite{radford2021learning})}
    \label{fig:appendix_tsne_e}
  \end{subfigure}\hfill
  \begin{subfigure}[t]{0.22\textwidth}
    \centering
    \includegraphics[width=\linewidth]{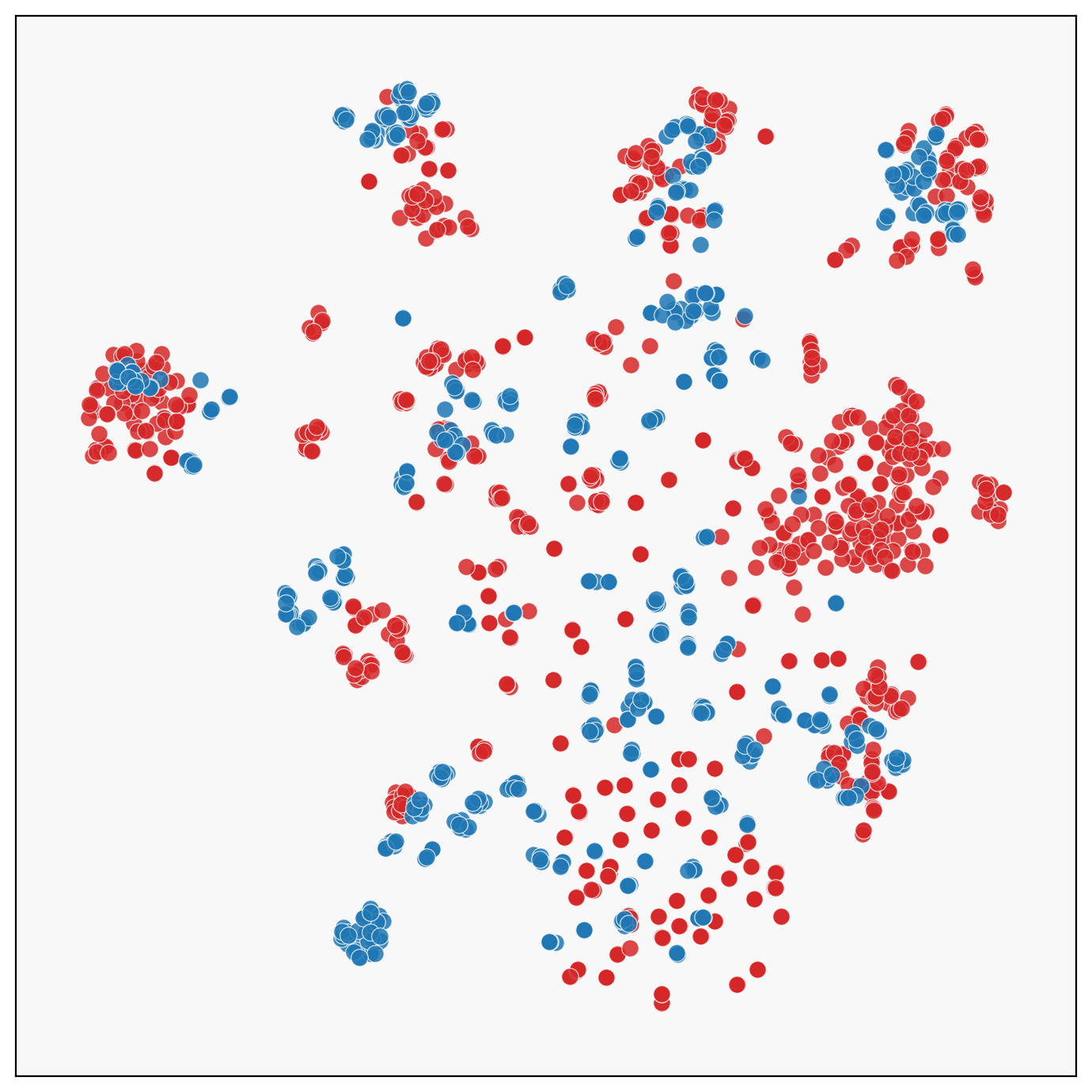}
    \caption{ViTTA~\cite{lin2023video}}
    \label{fig:appendix_tsne_f}
  \end{subfigure}\hfill
  \begin{subfigure}[t]{0.22\textwidth}
    \centering
    \includegraphics[width=\linewidth]{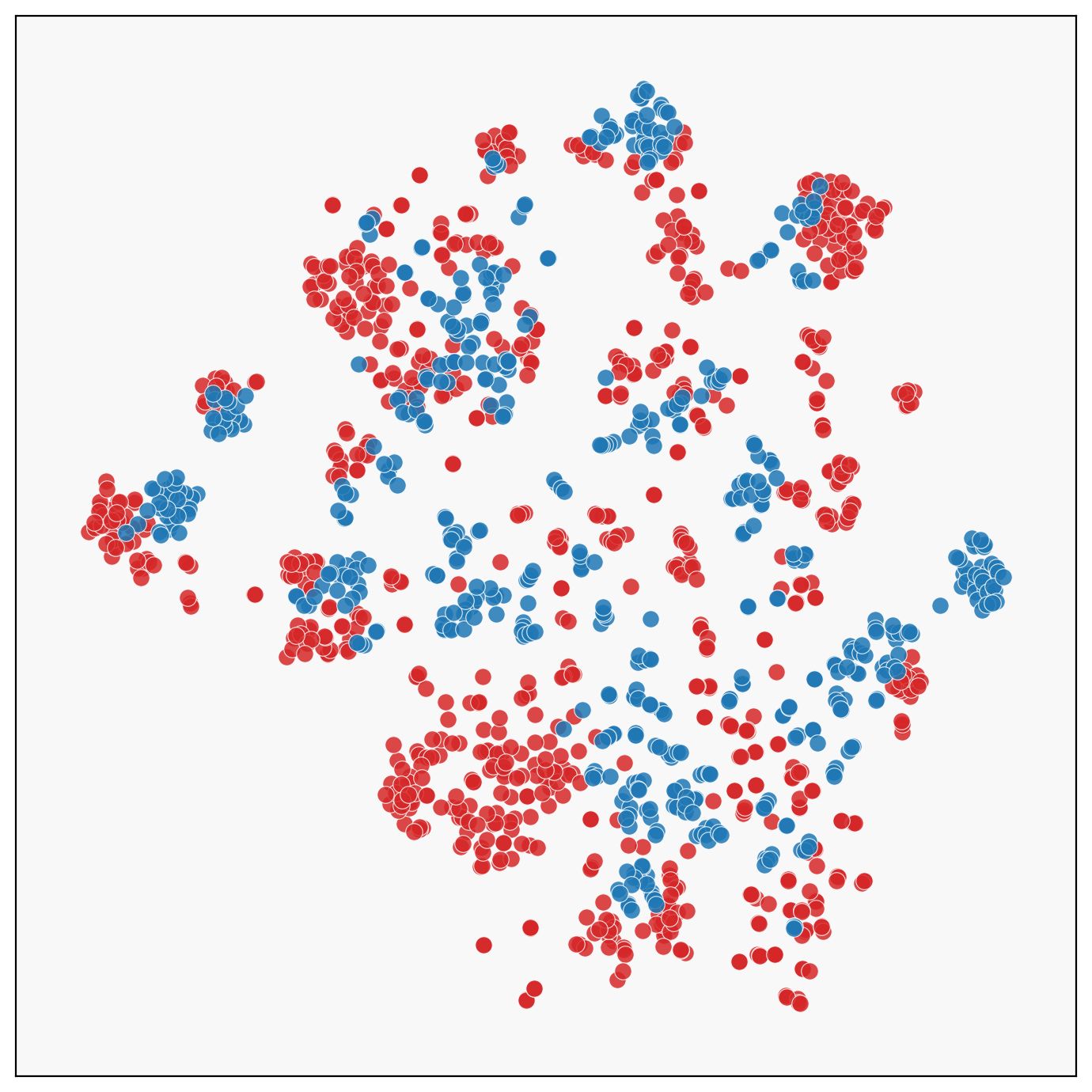}
    \caption{BATCLIP~\cite{maharana2025batclip}}
    \label{fig:appendix_tsne_g}
  \end{subfigure}\hfill
  \begin{subfigure}[t]{0.22\textwidth}
    \centering
    \includegraphics[width=\linewidth]{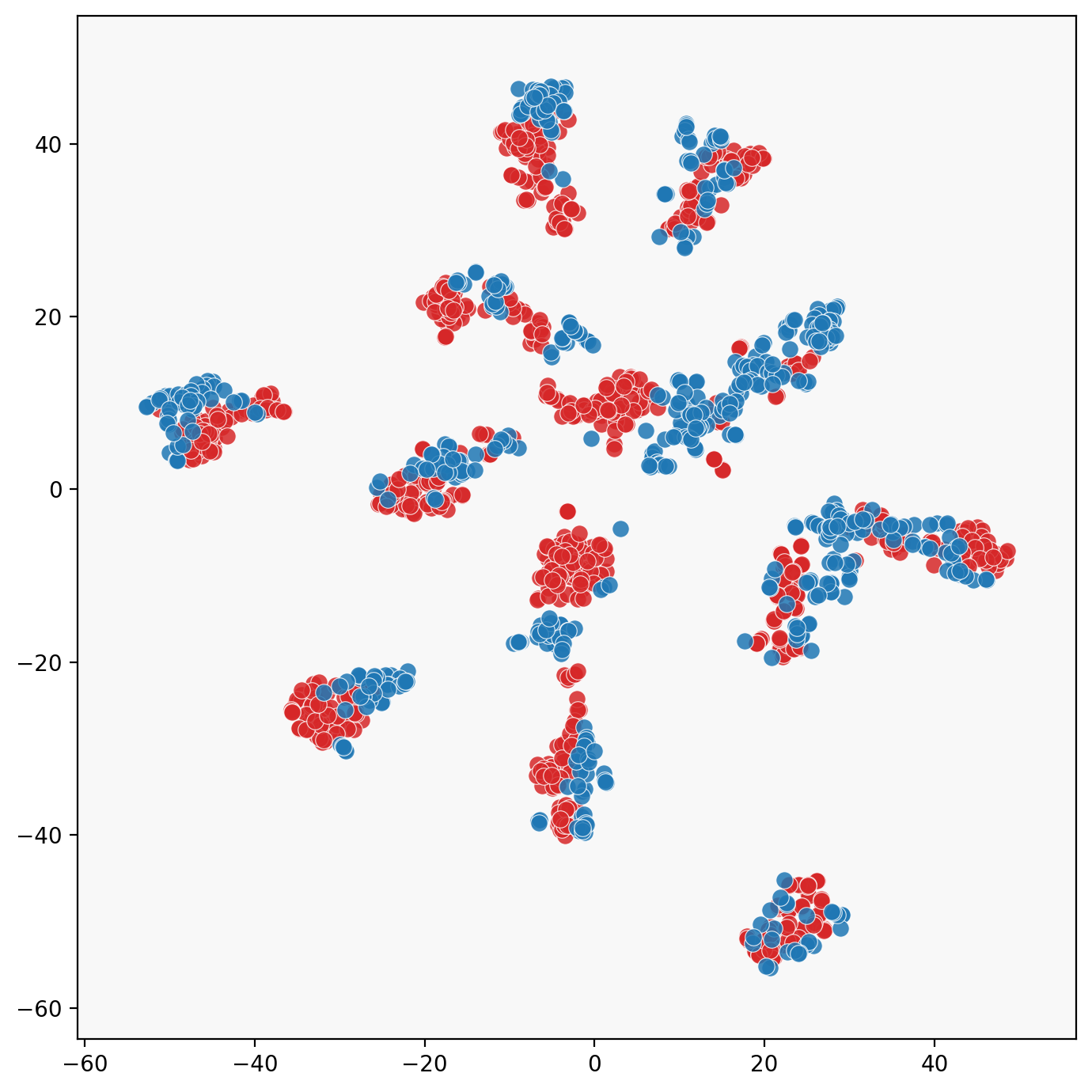}
    \caption{TADD (ours)}
    \label{fig:appendix_tsne_h}
  \end{subfigure}
  \caption{Visualization of features learned by each method on the HMDB51$\rightarrow$UCF101 target stream (settings reflected in the file names). Subpanels (a)--(d) are class-colored t-SNE (zero-shot CLIP, ViTTA, BATCLIP, TADD, left to right); (e)--(h) are domain-colored t-SNE for the same methods in the same order.}
  \label{fig:appendix_tsne}
\end{figure*}

Figure~\ref{fig:appendix_tsne} provides two-dimensional t-SNE projections of the video-level features extracted by each model on the HMDB51$\rightarrow$UCF101 benchmark.

The top row ((a)--(d)) colors points by action class to reflect how well each representation separates categories on the target domain.  Zero-shot CLIP (a) provides a reasonable semantic layout inherited from vision-language pre-training, but the clusters are not always compact, and some categories remain entangled because the visual encoder was never tuned on HMDB-style video statistics. Video TTA baselines like ViTTA (b) and BATCLIP (c) alter the feature geometry through test-time optimization. However, without a strong global anchor, adaptation stresses batch norms and auxiliary losses, which can introduce irregular and fractured layouts. In contrast, TADD (d) is trained to respect both CLIP's zero-shot predictions and the source adapter's discriminative structure. Qualitatively, this yields tighter, more class-cohesive neighborhoods than the baselines, directly aligning with the quantitative closed-set gains reported.

The bottom row ((e)–(h)) retains the source data but colors points by domain rather than by class, exposing domain structure independently of semantic grouping. Zero-shot CLIP (e) shows visible domain separation, as the backbone features still carry biases from large-scale pre-training. After adaptation, ViTTA (f) and BATCLIP (g) either preserve a substantial domain split or rearrange the embeddings in ways that fail to translate into cleaner class manifolds. TADD (h), however, successfully combines zero-shot regularization with source-knowledge distillation. Target updates reshape the adapter without drifting away from CLIP's domain-agnostic semantics. Consequently, within the class-consistent clusters, the domain labels remain comparatively mixed rather than forming a dominant axis of variation.

Taken together, these two rows reinforce the findings drawn from the quantitative results: effective video TTA must simultaneously sharpen class structure on the target stream while preventing domain identity from becoming an overly dominant factor in the learned embedding.

\end{document}